\documentclass[11pt]{article}
\usepackage{coling2020}
\usepackage{times}

\usepackage{soul}
\usepackage{url}
\usepackage[hidelinks]{hyperref}
\usepackage[utf8]{inputenc}
\usepackage[small]{caption}
\usepackage{graphicx}
\usepackage{amsmath}
\usepackage{verbatim} 
\usepackage{booktabs}
\usepackage{tikz}
\usepackage{amssymb}
\usepackage{pifont}
\usepackage[ruled,vlined]{algorithm2e}
\usepackage{array}
\usepackage{xcolor}
\usepackage{mathtools}
\usepackage{amsmath}
\usepackage[ruled,vlined]{algorithm2e}

\newenvironment{sysmatrix}[1]
{\left(\begin{array}{@{}#1@{}}}
	{\end{array}\right)}
\newcommand{\ro}[1]{%
	\xrightarrow{\mathmakebox[\rowidth]{#1}}%
}
\newlength{\rowidth}
\AtBeginDocument{\setlength{\rowidth}{3em}}

\colingfinalcopy 

\title{Exclusion and Inclusion - A model agnostic approach to feature importance in DNNs}

\author{
Subhadip Maji*\\
  Optum, UnitedHealth Group \\
  Bengalure, India\\
  {\tt maji.subhadip@optum.com} \\\And
  Arijit Ghosh Chowdhury* \\
  Optum, UnitedHealth Group \\
  Bengaluru India \\
  {\tt arijit.ghosh@optum.com}  \\\AND
    Raghav Bali* \\
  Optum, UnitedHealth Group \\
  Bengaluru India \\
  {\tt raghavbali@optum.com} \\\And
      Vamsi M Bhandaru \\
  Optum, UnitedHealth Group \\
  Hyderabad, India \\
  {\tt vamsi.bhandaru@optum.com} 
  \\}
\date{}

\begin{document}
\maketitle
\begin{abstract}
Deep Neural Networks in NLP have enabled systems to learn complex non-linear relationships. One of the major bottlenecks towards being able to use DNNs for real world applications is their characterization as black boxes. To solve this problem, we introduce a model agnostic algorithm which calculates phrase-wise importance of input features. We contend that our method is generalizable to a diverse set of tasks, by carrying out experiments for both Regression and Classification. We also observe that our approach is robust to outliers, implying that it only captures the essential aspects of the input.
\end{abstract}

\section{Introduction}

\textbf{Motivation:} Neural networks have rapidly become a central component in NLP systems in the last few years. The improvement in accuracy and performance brought by the introduction of neural networks has typically come at the cost of our understanding of the system. Deep learning systems have demonstrated superior performance, compared to traditional Machine Learning techniques. This is due to their ability to learn complex, non-linear, dependencies between features . However, the inability to effectively explain these dependencies has led to neural networks being characterized as black boxes . Moreover, the use of black-box models have come under scrutiny for lack of fairness and intepretability.
There has been growing body of work that deals with interpreting neural architectures \cite{ribeiro2016should,murdoch2018beyond,singh2018hierarchical}.  For Natural Language Processing, specifically in sequence-to-sequence tasks, there are relatively few methods that can extract the interactions between features that a DNN has learned.  There is a growing need for neural networks to be interpretable for deployment in real world applications.

\noindent \textbf{Our work} We introduce a model agnostic technique, \textit{Exclusion-Inclusion} for phrase wise feature importance in DNNs. We demonstrate our methodology on both classification and regression tasks.

Our work makes the following contributions :

\begin{itemize}

\item For classification tasks, we highlight which phrases contribute to a specific class.

\item For regression tasks, we calculate the positive/negative impact of phrases on the overall score.

\end{itemize}

\section{Related Work}
\label{intro}

There has been a growing body of work that deals with word level importance scores for deep learning models.  \cite{murdoch2018beyond} introduced a contextual decomposition of the LSTM’s output embedding into a sum over word coefficients, and demonstrated that those coefficients are meaningful by using them to distill LSTMs into rules-based classifiers. \cite{li2016understanding} introduced Leave One Out, a technique that observes the change in log probability resulting from replacing a given word vector with a zero vector. \cite{sundararajan2017axiomatic} proposed a gradient-based technique, called Integrated Gradients, which was evaluated anecdotally. These methods, however are limited to word wise importance and only work for classification.

There have been methods that calculate phrase level feature importance using Contextual Decomposition \cite{murdoch2018beyond} and Hierarchical Interpretations \cite{singh2018hierarchical}. However \newcite{murdoch2018beyond} only work with LSTMs and \newcite{singh2018hierarchical} do not show how a phrase affects every class present in the dataset. To our knowledge, there has been no previous work that deals with phrase level feature importance for regression tasks in NLP.
Attention based models \cite{bahdanau2014neural} offer another means of providing some interpretability. Such models have been successfully applied to many problems, leading to performance improvements \cite{strobelt2017lstmvis} . However, attention is at best an indirect indicator of importance, with no directionality. It doesn't tell us which class the phrase is important for, or in the case of regression, whether it increases/decreases the actual score. Attention weights. can be used to visualize which parts of a sentence are being focused on , but cannot be used as an interpretation technique.  Attention is also incapable of modelling interactions between words, hence we cannot calculate phrase wise attention.

%
%
\blfootnote{
    %
    %
    \hspace{-0.65cm}  
    Place licence statement here for the camera-ready version. See
    Section~\ref{licence} of the instructions for preparing a
    manuscript.
    %
    %
    %
    %
}

\section{Methodology}

\subsection {Problem Statement}

We are given a dataset $D = \{x_{ki}, y_{ki}\}_{i=0}^N$. Each $x_{ki}$ is string of text and label $y_i  \{1....q\}$ indicates the label/value of $x_{ki}$ among a given set of $q$ labels. We fit a model $F$ such that $y_{ki} = F(x_{ki})$.  Using the Exclusion-Inclusion algorithm, we determine word phrases that (i) contribute to a particular class (Classification) ii) Positively or negatively impact the score (Regression). 

\subsection { Exclusion and Inclusion }
This method calculates the local effects (positive, negative or neutral) of words (or phrases) with inclusion and  exclusion of that particular word (or phrase) for a data point. For regression while calculating the effects, there is one intermediate step i.e. importance of words (or phrases) calculation and regression losses (e.g. MAE, MSE, etc) are being used while doing so. To calculate the final effects, for regression we used the predicted response($\hat{y}$) and for classification we used predicted probability. 

\subsubsection{Importance Calculation}
This step is very important while effect calculation for regression, because after removing the unimportant words, we calculate the effect of only the important phrases on the output. We do not calculate the effect of any word (or phrase) which are unimportant. This motivation came from the significance of coefficients of linear regression. For the given linear regression Equation \ref{eq:regression}, the $\hat{\beta_i}$ values are the estimated coefficients. This sign of coefficients tells that whether $x_i$ is effecting the output ($\hat{y}$) positively or negatively and values tells the impact of the effect. But before deciding so, we perform the t-test on the $\beta_i$ to check whether they are significant or not. To perform the t-test we calculate the standard error of the $\beta_i$. As we assume that $\beta_i$ values follow t-distribution and we have one sample (the training data), it is easy to calculate the standard error as well as the p-value of $\beta_i$ to check it's significance (or importance). In NLP deep model we pass the input as sequence and the occurrence of the same word (or phrase) at the same location in the training data almost does not repeat. This means that the number of sample of that word (or phrase) in that location is always almost 1 which makes difficult to calculate the standard error of that word (or phrase) at that location. 
\begin{equation}
	\hat{y} = \hat{\beta_0} + \hat{\beta_1} x_1 + \hat{\beta_2} x_2 + ... + \hat{\beta_n} x_n
	\label{eq:regression}
\end{equation}

To tackle this problem we used the loss function of the problem as our pivot and checking the change of this loss function we decide whether that word (or phrase) is important or not. For a regression example, we calculated the predicted response excluding and including one phrase say, $phr_i = [w1,w2,..,wn]$ and got the output to be respectively, $\hat{y_{ex}}$ and $\hat{y_{in}}$. We calculate the losses (MAE or MSE for regression) for each of outputs i.e. $MAE_{ex} $ (or $MSE_{ex}$) and $MAE_{in} $ (or $MSE_{in}$). Now if $MAE_{in} < MAE_{ex}$, this means that inclusion of $phr_i$ reduces the loss. That means this phrase $phr_i$ is important to the output and if the reverse happens in the loss then that phrase is unimportant. It is seen that after removing all the unimportant phrases following the described procedure, if we predict the output only on the important words then the new output becomes more close with respect to the previous predicted output based on all the words. Mathematically, if including all the words if the loss is $MAE_{all}$ and including the important words the loss is $MAE_{imp}$ then always $MAE_{imp} <= MAE_{all}$.

The same procedure is not followed for classification, because for classification instead of direct score, we predict the probability of the data point to assign to a class. For classification, we calculate which words (or phrases) increase or decrease the probability to the data point to fall in that specific class. This thing has been captured in the effect calculation.

\subsubsection{Effect Calculation}
This section describes the calculation of effect of the word (or phrase) on the output Here, effect indicates the positive or negative impact of the word (or phrase) on the output i.e. for regression if any phrase $phr_j$ has positive effect that means inclusion of that particular phrase will increase the output response ($\hat{y}$) and exclusion of that word will decrease the output response analogous to $\hat{\beta}$ in regression. It is the other way around if that phrase has negative impact. For classification, positive impact of any word (or phrase) means inclusion of that word (or phrase) increase the probabilty of that data point to fall in that class and vice versa for negative impact. For multi-class classification we have performed the this effect analysis for each of the classes. This is described below.

Before calculating the effect we replace the identified unimportant words with zero. Zero means the absence of that word which is analogous to what we do during padding or masking. Calculation of effect is similar to the calculation of importance, but instead of loss function we take the predicted output of the model here. For regression, let's say including this phrase $phr_i = [w1,w2,..,wn]$, we get predicted response $\hat{y_{in}}$ and excluding this phrase we get predicted response to be $\hat{y_{ex}}$. If $\hat{y_{in}} > \hat{y_{ex}}$, this means that inclusion of that phrase increases the output of the model and exclusion of that phrase decreases the output response. That means this phrase has positive effect on output and vice version if the equation is reverse. For classification if there are n classes $\{c_1, c_2, ... , c_n\}$, we calculate the probability of each class for that data point including the phrase $phr_i = [w1,w2,..,wn]$ of interest and one more time excluding the same. Let's say, probabilities of class $j$, including and excluding $phr_i$ are $P(c_j)_{in}$  and $P(c_j)_{el}$ respectively. If $P(c_j)_{in} > P(c_j)_{el}$, this means that inclusion of that phrase increases the probability of the data point to fall in class $c_j$ and vice versa.

The flowchart of the above-mentioned process is shown in the Figure \ref{fig:flowchart}. For classification there is no importance calculation step, so the input will be directed to Effect calculation as the first step.

\subsubsection{Evaluation Metric}
The evaluation metric of effect is the traditional percentage change of model output with respect to excluding and including phrases. We call it EI (Exclusion-Inclusion) score, which is defined as,
\begin{equation}
	EI(phr_i) = \frac{\hat{y_{in}} - \hat{y_{el}}}{\hat{y_{in}}} * 100
\end{equation}
If the EI score is positive then, the phrase has positive effect and if negative then it has negative effect on the output. The above equation is shown for regression, for classification $\hat{y}$ can be simply replaced with $P(c_j)$ to get the effect of phrases on that specific class.

\subsubsection {Algorithm}
To build the algorithm, let us introduce the prerequisites. Assume, we have the trained model M, where M(x) is model output. This output is $\hat{y}$ for regression and $\hat{P(c_j)}$ for classification. We will calculate the effects of words (or phrases) on sentence S, and $S : \{w_1 w_2 w_3 ... w_n\}$ : [$i_{w_1}, i_{w_2}, i_{w_3}, ... , i_{w_n}$], where $w_j$ means the j-th word in S and $i_{w_j}$ is the index of j-th word in the vocabulary. 

\begin{algorithm}[H]
	\SetAlgoLined
	\KwIn{Input Sentence}
	\KwResult{Get phrases with positive, negative and neutral effect on output}
	\textbf{Part 1: Importance Calculation} \\
	\nl $k = 0$\;
	\nl \While{$k <= n$}{
		Calculate loss omitting $i_{w_k}$ = $loss_{(i_{w_k})}$\;
		Calculate loss omitting $i_{w_k}$ and $i_{w_{k+1}}$ = $loss_{(i_{w_k}, i_{w_{k+1}})}$\;
		\eIf{$loss_{(i_{w_k})} < loss_{(i_{w_k}, i_{w_{k+1}})}$}{
			$j = 1$ \\
			\While{$loss_{(i_{w_k}, i_{w_{k+1}}, ..., i_{w_{k+j}})} < loss_{(i_{w_k}, i_{w_{k+1}}, ..., i_{w_{k+j}}, i_{w_{k+j+1}})}$}{
			$j = j+1$}\
			Mark $[i_{w_k}, i_{w_{k+1}}, ... , i_{w_{k+j}}]$ as un-important\;
		}{
			$j = 1$ \\
			\While{$loss_{(i_{w_k}, i_{w_{k+1}}, ..., i_{w_{k+j}})} > loss_{(i_{w_k}, i_{w_{k+1}}, ..., i_{w_{k+j}}, i_{w_{k+j+1}})}$}{
				$j = j+1$}\
			Mark $[i_{w_k}, i_{w_{k+1}}, ... , i_{w_{k+j}}]$ as important\;
		}
	}
	\nl Filter out all the important word indices and replace the rest indices with 0 and get model predictions on the important words i.e. $\hat{y_{imp}}$ \\
	\textbf{Part 2: Effect Calculation} \\
	\nl $k = 0$\;
	\While{$k <= n$}{
		\For{j $\gets$ k to n}{
			Calculate EI score by excluding phrases [$i_{w_k}, i_{w_{k+1}}, ... ,i_{w_{j}}$], \\ 
			\begin{equation}
			EI_{(i_{w_k}, i_{w_{k+1}}, ... ,i_{w_{j}})} = \frac{\hat{y_{imp}} - \hat{y_{(i_{w_k}, i_{w_{k+1}}, ... ,i_{w_{j}})}}}{\hat{y_{imp}}} * 100
			\end{equation}
			
		}
	}
	\nl Conclude phrase with positive and negative EI scores will have positive and negative effect on output respectively. The un-important phrases (or words) are considered as neutral.
	\caption{Exclusion Inclusion Algorithm}
\end{algorithm}

While performing any kind of feature analysis, we do have the true output most of the times. If any case we do not have that true output or if we are performing classification then Part 1 of the algorithm to be skipped and only the Part 2 to be executed. 
\begin{figure}
	\centering
	\includegraphics[scale=0.75]{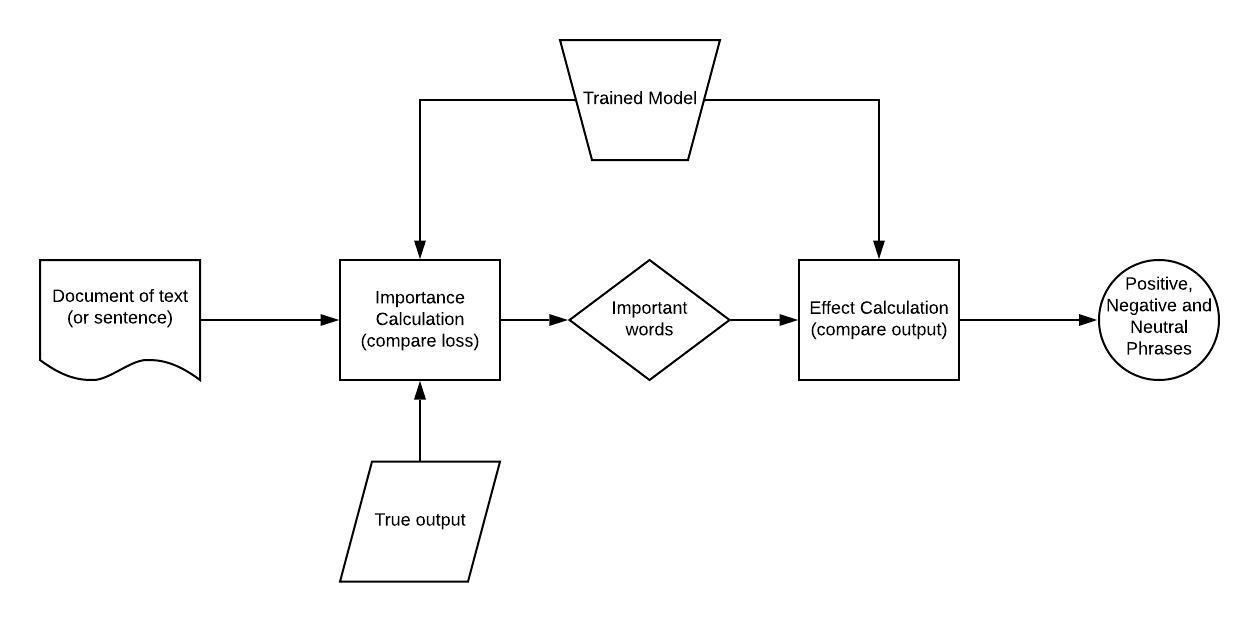}
	\caption{Flowchart of the Exclusion Inclusion method}
	\label{fig:flowchart}
\end{figure}

\subsubsection{Real Time Exclusion and Inclusion}
The algorithm described above captures the actual effect of the words (or phrases) because their effects have been calculated with the presence of the other words in the model. This is always required to calculate any feature importance. One thing we also see that the number of model predictions for a sentence of n words is two times (two times for regression, one time for classification) the sum of consecutive n terms from 1 to n i.e. $\frac{2n(n+1)}{2}$. Two time because one for importance calculation and another for effect calculation. If n increases then this number also increases rapidly which in turns make our algorithm slower to execute. To handle this issue we parallelized the algorithm by building matrices. To calculate uni-grams we built a $n\times n$ square matrix and made the diagonals zero, where n is length of sequence of sentence S. Doing so, previously to calculate model predictions without uni-grams the required time complexity was $O(n)$, now it comes down to $O(1)$ as we are predicting all at once. The same can be done for bi-grams, tri-grams and more also, but the main difference is that for bi-grams the diagonal and next to diagonal will be zero and the size of the matrix will be $(n-1) \times n$. For calculating the effects of g-grams, the size of the matrix will be $(n-g+1) \times n$. The formulation of the matrices is shown in the Figure \ref{matrix_uni_bi}. Following this parrallelization we can bring down the effect calculation time very significantly i.e. $2n$ instead of $\frac{2n(n+1)}{2}$.

\begin{figure}
	\centering
	\begin{minipage}{\linewidth}
		\begin{alignat*}{2}
			\begin{sysmatrix}{rrrrrr}
				i_{w_1}&i_{w_2}&i_{w_3}&i_{w_4} & ... & i_{w_n}\\
				i_{w_1}&i_{w_2}&i_{w_3}&i_{w_4} & ... & i_{w_n}\\
				i_{w_1}&i_{w_2}&i_{w_3}&i_{w_4} & ... & i_{w_n}\\
				... & ... & ... & ... & ... & ...\\
				... & ... & ... & ... & ... & ...\\
				i_{w_1}&i_{w_2}&i_{w_3}&i_{w_4} & ... & i_{w_n}
			\end{sysmatrix}
			&\!\begin{aligned}
				&\ro{unigram}
			\end{aligned}
			\begin{sysmatrix}{rrrrrr}
				0&i_{w_2}&i_{w_3}&i_{w_4} & ... & i_{w_n}\\
				i_{w_1}&0&i_{w_3}&i_{w_4} & ... & i_{w_n}\\
				i_{w_1}&i_{w_2}&0&i_{w_4} & ... & i_{w_n}\\
				... & ... & ... & ... & ... & ...\\
				... & ... & ... & ... & ... & ...\\
				i_{w_1}&i_{w_2}&i_{w_3}&i_{w_4} & ... & 0
			\end{sysmatrix}
		\end{alignat*}
		\centering (a)
		
		\begin{alignat*}{2}
		\begin{sysmatrix}{rrrrrr}
		i_{w_1}&i_{w_2}&i_{w_3}&i_{w_4} & ... & i_{w_n}\\
		i_{w_1}&i_{w_2}&i_{w_3}&i_{w_4} & ... & i_{w_n}\\
		i_{w_1}&i_{w_2}&i_{w_3}&i_{w_4} & ... & i_{w_n}\\
		... & ... & ... & ... & ... & ...\\
		... & ... & ... & ... & ... & ...\\
		i_{w_1}&i_{w_2}&i_{w_3}&i_{w_4} & ... & i_{w_n}
		\end{sysmatrix}
		&\!\begin{aligned}
		&\ro{bigram}
		\end{aligned}
		\begin{sysmatrix}{rrrrrr}
		0&0&i_{w_3}&i_{w_4} & ... & i_{w_n}\\
		i_{w_1}&0&0&i_{w_4} & ... & i_{w_n}\\
		i_{w_1}&i_{w_2}&0&0 & ... & i_{w_n}\\
		... & ... & ... & ... & ... & ...\\
		... & ... & ... & ... & ... & ...\\
		i_{w_1}&i_{w_2}&i_{w_3}&i_{w_4} & ... & 0
		\end{sysmatrix}
		\end{alignat*}
		\centering (b)
	\end{minipage}
\caption{Formulation of matrices for uni-grams and bi-grams}
\label{matrix_uni_bi}
\end{figure}

\subsubsection{Issue Handling with very long sequences}
Though with parallelization the time required for EI methods comes down significantly, it is still a problem where sequences are very long (e.g. essay, etc.) i.e. $n$ is large. In that case the number of model predictions will still be $2n$, but due to the large matrix sizes the prediction time will be slower. To handle that issue, we tried to reduce the row number of matrices as the number of grams increase, as a result the second part of the algorithm where we calculate effect will change slightly. Initially, for sentence S, we started iteration from $w_i$ and would continue the iteration till phrase $\{w_i w_{i+1} w_{i+2}...w_n\}$. But now instead of looking for all the possible combination we will introduced one early iteration stop criteria as we did in the importance calculation section. Starting from $w_i$, we calculated the predicted output omitting that word, if the predicted output decreases, then we marked $w_i$ as the word having positive effect and now omitted phrase $\{w_i w_{i+1}\}$, if the output decreases more, we continue the iteration until the predicted output is not lesser than the previous one which means $\hat{y_{(i, i+1,i+2,...,j)}} > \hat{y_{(i, i+1,i+2,...,j-1)}}$ where $w_j$ is the word where the iteration terminates. In that case, phrase $\{w_i w_{i+1} ... w_{j-1}\}$ will be the phrase with positive effect on output. Same iteration condition is applicable for the phrase with negative effect also. 

Doing so the number of iteration reduces significantly as the number of gram increases which in turns reduces the number of row in the matrices making model predictions faster. It is seen that generally the iteration happens up to 10 grams for sequence with length of 5000 words.

\section{Experiments}

We perform our experiments using an LSTM \cite{hochreiter1997long} for both regression and classification experiments. We use 3 LSTM layers each of hidden-size 128. We use 300-dimensional glove embeddings with trainable parameters.
For all models we use the Adam \cite{kingma2014adam} optimizer with learning rate of $2e-5$. For classification, we use Cross Entropy loss \cite{zhang2018generalized} and for regression, we use Mean Squared Error \cite{wang2009mean}.

We use 3 datasets for evaluating our methodology:
 \begin{itemize}
     \item \textbf{SST-2 \cite{socher2013recursive}}: The Stanford Sentiment Treebank has movie reviews divided into 2 categories, negative and positive. This is a classification problem.
     \item \textbf{ASAP} \footnote{https://www.kaggle.com/c/asap-aes}: This dataset has a total of 7k essays for Automatic Essay Scoring. We treat this as a regression problem. 
    
 \end{itemize}

\section{Observations}

\textbf{Classification: } Table $1$ shows 5 examples from the SST-2 dataset. The effects are calculated for the true class. The phrases contributing to the positive class are annotated as green while phrases which contribute to the negative class are annotated as red. We highlight how one sentence can have multiple phrases contributing to either class ( For example, \textit{ bad movie} and \textit{good actors} ).
It also highlights possible reasons for the mis-classified samples. For example, \textit{There's no art here}, supersedes any positive phrase within the sentence, although the conclusion itself is positive. It is interesting to note that the contribution of one very negative phrase can cancel out contributions from multiple positive phrases. In cases where there is no ambiguity, the entire sentence contributes to one class unanimously ( For example, \textit{A fun ride}). Through these examples, we elucidate how the Exclusion-Inclusion method adds phrase wise interpretability to DNNs.

\begin{center}
  \begin{table*}[h]
  \renewcommand*{\arraystretch}{1.4}
\centering
\begin{tabular}{c c c c}
\hline
\textbf{Sentence} & \textbf{Predicted Class} & \textbf{Label}  \\
\noalign{\smallskip}
\hline

\colorbox{red}{a bad movie} that happened to \colorbox{green}{good actors} & 0 & 0 \\
it's \colorbox{red}{simply stupid irrelevant} and \colorbox{red}{deeply truly bottomlessly cynical} & 0 & 0 \\
\colorbox{red}{there s no art here} it s \colorbox{green}{still a good yarn }which is \colorbox{green}{nothing to sneeze at} these days & 0 & 1 \\
 \colorbox{green}{a fun ride} & 1 & 1 \\
 overcomes its \colorbox{red}{visual hideousness} with a \colorbox{green}{sharp script and strong performances} & 1 & 1 \\

\hline
\end{tabular}
\caption{Sample interpretations from SST-2}
\label{tab:dataset}     
\end{table*}
\end{center}

\begin{figure}
	\centering
	\includegraphics[scale=0.75]{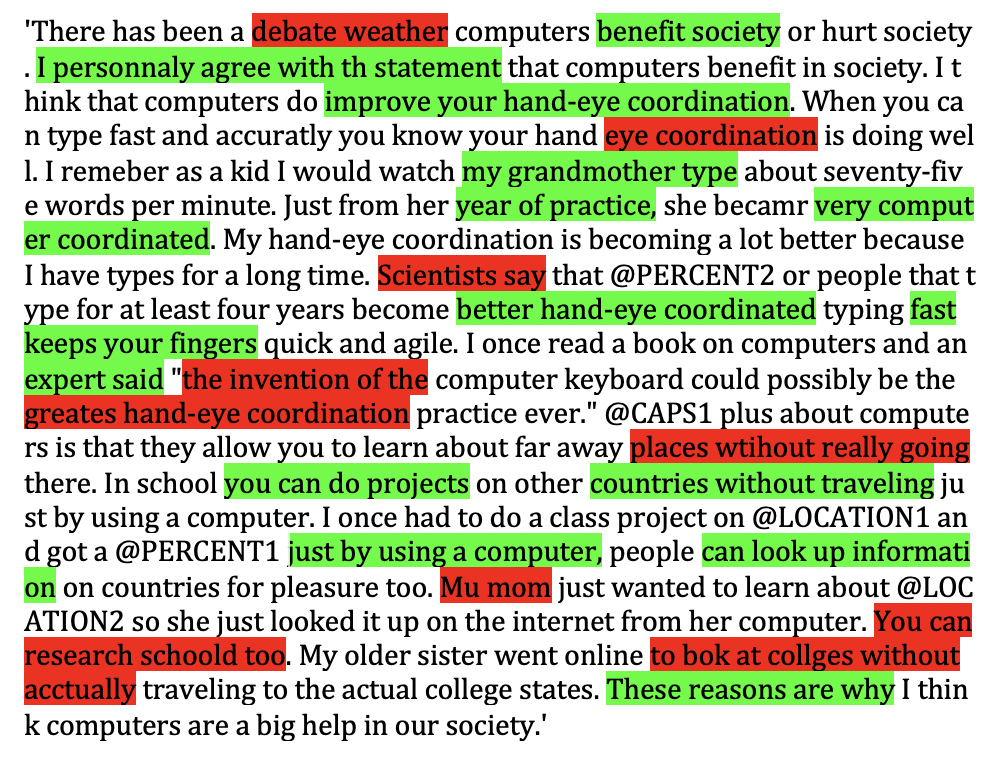}
	\caption{Sample enablers and disablers for regression}
	\label{fig:flowchart}
\end{figure}

\noindent \textbf{Regression: }  Figure 3 shows a sample essay from Automatic Essay Scoring task on the ASAP dataset. The phrases contributing to a higher score are annotated as green while phrases which decrease the score are annotated as red. We note that spelling mistakes are marked as red because they are often out-of-vocabulary. Furthermore, descriptive phrases are marked positive. Phrases which keep repeating themselves are marked negative after multiple occurrences ( For example \textit{Hand Eye Coordination}. We have deliberately not used any handcrafted features for AES, as used in relevant literature. We contend that interpretable predictions from a simple LSTM model will aid in creating better handcrafted features for Automatic Essay Scoring, and other general regression tasks.

\section{Conclusion and Future Work}

In this paper, we introduce Exclusion-Inclusion, a model agnostic method for phrase wise feature importance in DNNs. Through qualitative experiments, we demonstrate how our approach adds an extra layer of interpretability for common classification and regression tasks. We also highlight how our method can be applied in a real world setting with very long sequences, where compute resources are limited. Our work is a step towards better understanding of how DNNs predict text, as opposed to treating complex non linear models as black boxes.

\nocite{*}
\bibliographystyle{coling}
\bibliography{coling2020}

\end{document}